\pdfoutput=1
\documentclass[11pt]{article}

\usepackage[]{EMNLP2023}

\usepackage{times}
\usepackage{latexsym}

\usepackage[T1]{fontenc}
\usepackage[utf8]{inputenc}

\usepackage{microtype}

\usepackage{inconsolata}

\usepackage{subcaption}
\usepackage{graphicx}
\usepackage{amsmath}
\usepackage{multirow}
\usepackage{caption}
\usepackage{epsfig}
\usepackage{xcolor}
\usepackage{CJKutf8}
\usepackage{enumitem}
\usepackage{booktabs}

\title{Towards Safer Operations: An Expert-involved Dataset of High-Pressure Gas Incidents for Preventing Future Failures}

\author{Shumpei Inoue$^1$, Minh-Tien Nguyen$^{1, 2,}$\thanks{$^*$Corresponding Author.} , Hiroki Mizokuchi$^1$, Tuan-Anh D. Nguyen$^1$, \\\textbf{Huu-Hiep Nguyen}$^1$, \textbf{Dung Tien Le}$^1$ \\
        $^1$Cinnamon AI, 10th floor, Geleximco building, 36 Hoang Cau, Dong Da, Hanoi, Vietnam. \\
        \texttt{\{sinoue, ryan.nguyen, hmizokuchi, tadashi, hubert, nathan\}@cinnamon.is} \\
        $^2$Hung Yen University of Technology and Education, Hung Yen, Vietnam. \\
        \texttt{tiennm@utehy.edu.vn}}

\begin{document}
\maketitle
\begin{abstract}
This paper introduces a new IncidentAI dataset for safety prevention. Different from prior corpora that usually contain a single task, our dataset comprises three tasks: named entity recognition, cause-effect extraction, and information retrieval. The dataset is annotated by domain experts who have at least six years of
practical experience as high-pressure gas conservation managers.
We validate the contribution of the dataset in the scenario of safety prevention. Preliminary results on the three tasks show that NLP techniques are beneficial for analyzing incident reports to prevent future failures.
The dataset facilitates future research in NLP and incident management communities. The access to the dataset is also provided.\footnote{The IncidentAI dataset is available at: \url{https://github.com/Cinnamon/incident-ai-dataset}}

\end{abstract}

\section{Introduction}\label{sec:intro}
Daily activities usually face incidents that can significantly affect risk management.
In specific industries such as manufacturing, an incident can make a significant consequence that not only reduces the reputation of companies but also breaks the product chain and costs a lot of money.
It motivates the introduction of the safety-critical area where AI solutions have been proposed to prevent repeated failures from historical samples \cite{yampolskiy2019predicting,Sean-database-AAAI-21,durso2022analyzing,nor2022abnormality,chandra2023aviation,tikayat2023aerobert,andrade2023safeaerobert}.

%The solutions include the development of intelligent systems that utilize artificial intelligence for wide range purposes such as transportation \cite{national2017collision}, law enforcement \cite{dressel2018accuracy}, LegalAI \cite{zhong2020does}, and incident databases\footnote{https://cve.mitre.org}\footnote{https://www.faa.gov/data\_research/accident\_incident/} \cite{Sean-database-AAAI-21}.

There still exists a gap in the adoption of AI techniques for actual incident management scenarios due to the lack of high-quality annotated datasets.
The main challenges arise from two main reasons.
First, data annotation of incidents for AI-related tasks is a labor-expensive and time-consuming task that requires domain experts who have a deep understanding and excellent experience in their daily work.
Second, the collection of historical incidents is also challenging due to its dependence on the policies of companies.
We argue that the growth of the safety-critical area can be leveraged by introducing annotated incident datasets.

%the availability of infrastructure
%Finally, 
%As a result, a lot of studies only collect historic incidents for database creation \cite{Sean-database-AAAI-21}.

%because gas is the world's primary fuel sources cause the safety prevention problem and
%After careful discussion with domain experts and collection of common patterns in real projects, we show the scenario of IncidentAI for high pressure gas incidents in 

\begin{figure*}[!h]
    \centering
    \includegraphics[width=0.9\textwidth]{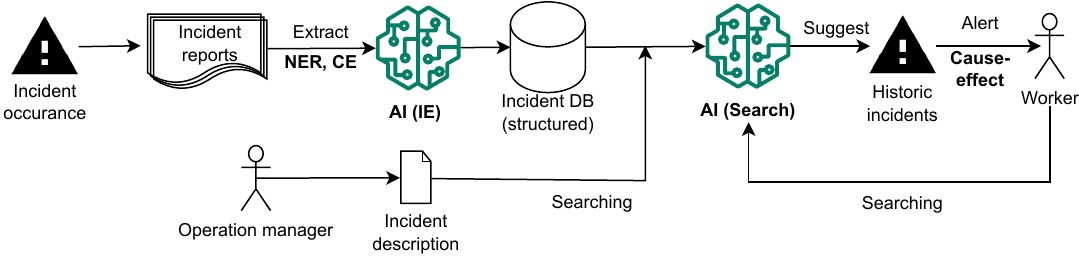}
    \caption{The scenario of IncidentAI in actual business cases. CE stands for cause-effect extraction.}
    \label{fig:scenario}
\end{figure*}

To fill the gap, this paper takes the high-pressure gas domain, a sector of the gas industry, as a case study.
This is because gas and its products are the major industry in the energy market that play an influential role in the global economy \cite{mokhatab2018handbook,pellegrini2019handbook}.
In addition, a gas incident may cost a lot of money with significant consequences.
The detection and analysis of past incidents are crucial for improving safety prevention and avoiding future failures.
Figure \ref{fig:scenario} shows the scenario of the detection and analysis.
The description of an occurred incident is noted in an incident report.
Then, important information (named entity and cause-effect extraction) from the incident report is extracted and stored in an incident database.
In operation, given the description of an incident, the manager can search historical incidents for potential risk analyses.
The system alerts the worker by showing historically relevant incidents and cause-effect information based on assigned tasks.
The worker can use suggested information for failure analysis to avoid future incidents.
In practice, given an incident report, workers, managers, or analyzers would like to know: (i) which aspects (entities) are relevant to the incident?, (ii) what is the cause and effect of the incident?, (iii) and which are historically relevant incidents of the current incident for risk and failure analyses?

%However, the retrieval and analysis of relevant incidents can be a non-trivial and time-consuming task and challenging for human analyses.
%It, therefore, raises a question that \textbf{whether AI solutions can be developed to support the analysis and management of gas incidents for the prevention of future failures?}

%For example, when an incident happens, analyzers would like to know the product of high-gas, the chemical of the product, or the process of the product.
% This information serves for later processing steps, e.g., IR.
%This extracted information is very meaningful for the analyzers to know more detailed information of the incident.
% The IR task is usually used to retrieve historic incidents for the current incident.
%By retrieving, analyzers can know the relevancy between the current and historic incidents.

To address the aforementioned questions, this paper introduces a new Japanese dataset that focuses on high-gas incidents and demonstrates the potential NLP applications in analyzing high-gas incident reports.
To do that, we first work closely with business members and domain experts to identify three potential NLP tasks: named entity recognition (NER), cause-effect extraction (CE), and information retrieval (IR) based on actual scenarios.
The NER task allows analyzers to extract fundamental units of an incident in the form of entities, e.g., the product or the process of the product.
This information is used to visualize statistics concerning key entities from past incidents retrieved through IR steps.
The CE task allows analyzers to extract the cause and effect of an incident.
The IR task is typically used to examine historical incidents similar to the current one and to develop countermeasures to prevent the recurrence of such incidents.
In business scenarios, information from the three tasks is vital for safety-critical and risk management.
This paper makes three main contributions as follows.
\begin{itemize}
    \item It introduces a new IncidentAI dataset that focuses on high-gas incidents for NER, CE, and IR. To the best of our knowledge, this is the first Japanese dataset that covers all three tasks in the context of high-gas incidents. It is annotated by domain experts to ensure a high-quality dataset that can assist in the efficient analysis of incident reports using AI models.
    %Although the size of the dataset is relatively small, i
    %\footnote{The dataset: \url{https://shorturl.at/iosxY}}

    \item It shows a scenario of IncidentAI in actual business cases. The scenario can serve as a reference for AI companies that are also interested in the analysis of incident reports.

    %\item It provides annotation guidelines that serve the purpose of maintaining consistency and reproducibility in the dataset annotations. These guidelines offer comprehensive instructions, criteria, methodologies, and standards to be adhered to during the annotation process. By sharing these guidelines, we aim to improve the overall quality and reliability of the dataset, enabling researchers and practitioners to obtain more precise and comparable results in their analyses and experiments.

    \item It benchmarks the results of AI models on NER, CE, and IR tasks that facilitate future studies in NLP and safety prevention areas.
\end{itemize}

%Specifically, we aim to evaluate the performance of AI models on the three tasks: named entity recognition (NER), cause-effect extraction (CE), and information retrieval (IR) and provide a baseline for future research in the NLP community. In addition, we provide annotation guidelines that serve the purpose of maintaining consistency and reproducibility in the dataset annotations. These guidelines offer comprehensive instructions, criteria, methodologies, and standards to be adhered to during the annotation process. By sharing these guidelines, we aim to improve the overall quality and reliability of the dataset, enabling researchers and practitioners to obtain more precise and comparable results in their analyses and experiments.

\section{Related Work}
\paragraph{Incident databases}
There exist industry-specific incident databases in many industries \cite{national2017collision}.
The databases contain a wide range of incidents such as cyber-security vulnerabilities and exposures \cite{MITRE}, aviation reports that contain accident and incident information of air flights \cite{Federal}, or reports of the pharmaceutical industry, healthcare providers, and consumers \cite{USDrug}.
%For example, the MITRE Corporation provides a database that contains more than 141K publicly disclosed cyber-security vulnerabilities and exposures \cite{MITRE}.
%a computer a cyber-security vulnerabilities database that allows users to search relevant information of computer security \cite{MITRE}.
%It contains more than 141K publicly disclosed cyber-security vulnerabilities and exposures.
%Federal Aviation Administration and National Aeronautics and Space Administration provides aviation reports that contain accident and incident information of air flights \cite{Federal}.
%United States Food and Drug Administration \cite{USDrug} has built a system that allows users to search for information related to the pharmaceutical industry, healthcare providers, and consumers.
%\footnote{https://incidentdatabase.ai}
Recently, an AIID database (AI Incident Database) was introduced \cite{Sean-database-AAAI-21}.
It indexes more than 1,000 publicly available incident reports.
%in several document types such as popular, trade, and academic press.
%The High Pressure Gas Safety Institute of Japan provides 18,171 high pressure gas incidents \cite{JPHPGS}.
%The dataset contains more than 16K incidents collected from the past.
%We extend the direction of collecting and building the corpora of IncidentAI. However, instead of only collecting incidents to create databases, we go further by creating a new Japanese dataset that composes three NLP tasks: NER, CE, and IR. It facilitate the analysis of high pressure gas incidents in a low-resource language.
The existing databases allow the storage of incident information, yet simple AI techniques (simple matching or classification) create a gap to analyze the incidents.
We leverage safety operations by introducing a new dataset that includes three main tasks: NER, cause-effect extraction, and IR. It facilitates the adoption of AI to prevent future failures in the context of high-pressure gas incidents.

\paragraph{NLP techniques}
have been applied to analyze incident reports \cite{Sean-database-AAAI-21,McGregor-indexing-AI-risks-22,Pittaras-taxonomy-system-cause-analysis-22,hong2021planning,nor2022abnormality,macrae2022learning,durso2022analyzing,shrishak2023deal,nor2022abnormality}.
The methods range from indexing for incident databases \cite{Sean-database-AAAI-21,McGregor-indexing-AI-risks-22,Pittaras-taxonomy-system-cause-analysis-22} to deeper analyses using machine learning \cite{durso2022analyzing,nor2022abnormality,chandra2023aviation}.
%For example, \citeauthor{Sean-database-AAAI-21} \citeyear{Sean-database-AAAI-21} developed an incident database that offers search a function for retrieving historic incidents.
%\citeauthor{durso2022analyzing} \citeyear{durso2022analyzing} analyzed failures stored by National Vulnerability Database\footnote{https://nvd.nist.gov} to  understand vulnerability trends, their root causes, and how to prevent future failures.
%\citeauthor{nor2022abnormality} \citeyear{nor2022abnormality} used Bayesian networks to for abnormality detection and failure prediction of real-world gas turbine.
Recently, BERT \cite{devlin2019bert} has been adapted to aviation safety \cite{chandra2023aviation,tikayat2023aerobert,andrade2023safeaerobert,jing2023bert}.
%\citeauthor{chandra2023aviation} \citeyear{chandra2023aviation} introduced Aviation-BERT trained by using accident and incident text narratives from the National Transportation Safety Board\footnote{https://www.ntsb.gov/Pages/home.aspx} (NTSB) and Aviation Safety Reporting System\footnote{https://asrs.arc.nasa.gov} (ASRS).
The recent survey also shows the role of NLP in aviation safety \cite{yang2023natural}.
%We extend IncidentAI by creating a new Japanese dataset that composes three NLP tasks: NER, CE, and IR. It facilitate the analysis of high pressure gas incidents in a low-resource language.
We share the direction of using NLP techniques in the analysis of incident reports. However, instead of focusing on single tasks, we are interested in three different tasks: NER, CE, and IR, that provide critical information for safety prevention. In addition, we provide a Japanese dataset to facilitate the creation of AI pipelines in a low-resource language.

\section{The HPGIncident Dataset}

\subsection{Data Collection}\label{subsec:data-collection}
The original dataset was collected from publicly available reports of high-gas incidents published in 2022 by the High-Pressure Gas Safety Institute of Japan.\footnote{\url{https://shorturl.at/BLWX6}} The original data contains descriptions of incidents, types of incidents, dates of incidents, industries, etc.
From the original 18,171 incident cases, 2,159 cases belonging to three industries: ``general chemistry", ``petrochemical", and ``oil refining" were first extracted.
These cases were used for the annotation of IR.
Subsequently, we selected 970 cases from that 2,159 cases based on the most recent dates for both the annotation of NER and CE tasks. We used the description of incidents as the input for annotation shown in the next section.

%From the original 18,171 incident cases, incidents in three industries: ``general chemistry", ``petrochemical", and ``oil refining" were first extracted, resulting to a total of 2,172 cases.

%of general chemistry, petrochemical, and oil refining respectively,

%at high-pressure gas plants

%https://www.khk.or.jp/public_information/incident_investigation/hpg_incident/incident_db.html

%SHUMPEI: PLEASE ALSO REVIEW and WRITE THE IR PART!

% The selection was carefully done by consulting domain experts to select three industries: ``general chemistry", ``petrochemical", and ``oil refining" based on its importance. 970 incident reports were chosen in descending order of the latest occurrence date.

\subsection{The Annotation Process}\label{subsec:data-annotation}
The dataset was created by three Japanese domain experts, each with at least six years of practical experience as high-pressure gas conservation managers. These experts possess qualifications as high-pressure gas production safety managers, a national certification demonstrating a certain level of knowledge and experience necessary to ensure the safety of high-pressure gas manufacturing facilities.

The process was divided into two steps: the creation of the guideline and the annotation of the entire dataset. In the first step, we randomly selected 100 samples from 970 collected samples for NER and CE, and from 2,159 collected samples for IR. Our team collaborated closely with experts to establish criteria for consistent annotations, including identifying the information types (entities) and their definitions for NER and CE, and determining the attributes that characterize incidents for IR. These criteria formed the basis of our guidelines. This initial stage was iterative, conducted in several rounds until a certain agreement score was achieved among the experts. This process played a vital role in training the annotators, ensuring that they shared a uniform understanding of the guidelines.
Once a high agreement score had been achieved, the remaining samples were apportioned into three segments, each corresponding to an annotator, who then proceeded to annotate their respective parts.
Subsequently, 100 random samples were selected from one annotator's portion. The other two annotators were tasked with annotating these 100 samples. For each task, NER, CE, and IR, an inter-annotator agreement was computed using these 100 samples. 
Due to space constraints, please refer to Appendix \ref{apx:appendix} for a more detailed explanation of annotation.
%The created dataset is small but very high quality annotated by domain experts. It costs one million Japanese Yen (around 7,170 USD) in five months.

% This is because we would like to keep the high quality of the dataset.
%and the annotation of high-gas incidents is much more challenging than other common domains.

%The BRAT\footnote{\url{https://brat.nlplab.org}} tool was used for the annotation of NER and CE while an excel file was used for the annotation of IR.

%The collected incident reports were annotated by domain experts in high-gas incidents with at least six years of experience in the field. The annotation was carried out using the guidelines developed by the research team. The guidelines were designed to ensure consistency and accuracy in the annotation process and were refined through an iterative process of discussion and feedback.

\paragraph{NER annotation}\label{para:NER}
As mentioned, entities provide basic information about an incident.  This represents the first tier in the incident report analysis. The initial step of NER annotation involves identifying the set of entities. The identification was carried out through meticulous coordination and several meetings with domain experts. Built on their insights and experiences, six critical entity types for incident analysis were established: \textbf{Products}, \textbf{Chemicals}, \textbf{Storage}, \textbf{Incidents}, \textbf{Processes}, and \textbf{Tests}. Table \ref{table:ner-entities} shows the definition of entities.
%and examples of the entities.
%which can be used to facilitate the first analysis of an incident.

\begin{table}[!h]
\centering
\setlength{\tabcolsep}{4pt}
\tiny
\caption{The definition of entities.}\label{table:ner-entities}
\begin{tabular}{lp{6.2cm}} \hline
Entity & Definition \\ \hline
Products & A noun phrase that mentions various gases.
Gaseous state at normal temperature and pressure. Do not tag items that are not general (things that do not appear even if you search the Web). Examples: mixed gas; flammable gas; refrigerant gas. \\ \hline

Chemicals & A noun phrase mentions chemical substances, reactants, and materials (other than gases) used in gas generation and process management. Items not included in the above Products. Examples: Benzene; Hydrocarbons. \\ \hline

Storage & General equipment where above
Products and Chemicals come into
contact. (i) Include equipment such as supports and insulators.
(ii) Include expressions that indicate the entire plant or facility.
(iii) Do not include expressions indicating parts such as entrances and exits if they are placed at the end of a word. Examples: tank; maturation furnace; refining tower; dehumidification tower. \\ \hline

Incidents & A phrase mentions incidents that resulted in or caused an accident, regardless of severity. It includes only incidents that actually occurred, and do not include situations that did not lead to an incident. Examples: seepage; leakage; fire; serious injury; death. \\ \hline

Processes & A phrase mentions handling of gas, and unit operations related to gas. Abnormal processes are included in incidents. Examples: filling; distillation. \\ \hline

Tests & A phrase mention inspection devices and inspection actions outside the production process line. Do not include inspection items such as XX concentration. Examples: inspection; visual inspection; leak test. \\

\hline
\end{tabular}\vspace{-0.2cm}
\end{table}

%, std: 0.045; (std: 0.009)
%(0.372 of F-1, std: 0.045)\footnote{https://github.com/kldtz/bratiaa} \cite{hripcsak2005agreement,kolditz2019annotating}.
%Several meetings were conducted to align with the annotators and update the guideline. After several rounds, the agreement score was 0.901 (std: 0.006).
%Second, after annotating the whole NER dataset, the agreement score of randomly cross-checking 100 other samples was lower than the score of 100 first samples (0.876 vs. 0.901).
%This is because there exist different opinions of annotators for 870 samples, but the score is good enough.

The annotation of NER uses the definition of entities in Table \ref{table:ner-entities} and the rules mentioned in the Appendix \ref{apx:ner-rules}. We observed two important points.
Firstly, the annotation was challenging even with domain experts.
In the first round of guideline creation, the agreement score was so low.
After several meetings, the agreement score was significantly improved.
It provides strong evidence for the adaptation of the whole NER dataset.
After annotating the whole NER dataset, the Fleiss' Kappa score of randomly cross-checking 100 other samples was 0.814, showing good agreement.
Secondly, entities are nested. It comes from the nature of data, for example, a product can contain a chemical or a process can include storage.
An annotated example of NER is shown in Figure \ref{fig:ner-sample} in the Appendix \ref{apx:ner-rules}.
Table \ref{tab:ner} summarizes the statistics of the NER dataset that follows the BIO format.
\begin{table}[!h]
\centering
\caption{Statistics of the NER dataset.}
\label{tab:ner}
\begin{tabular}{lccc} \hline
Entity & Train & Test & All\\
\hline
Product & 2,189 &  905 & 3,094 \\
Chemical & 1,440 &  576 & 2,016 \\
Storage & 5,881  & 2,251 & 8,132 \\
Incident & 5,274  & 2,045 & 7,319 \\
Process & 1,138 &  426 & 1,564 \\
Test & 1,615 & 572  & 2,187 \\
 \hline
\#entities in total & 17,537 &  6,775 & 24,312 \\
\#reports in total & 700 &  270 & 970 \\ \hline
\end{tabular}\vspace{-0.2cm}
\end{table}

\paragraph{Cause-effect annotation}\label{para:CE}
Causes and effects provide critical information about a given incident for the analysis, in which causes contain information about the cause of an incident and effects mention the consequences of the causes.
Similar to NER, we engaged in detailed discussions with domain experts to identify cause and effect types.
We observed that the cause is quite easy to identify while the effect composes several types such as the leakage of gas, physical damages (explosion or fire), human injuries caused by the incident, or others (not related to leakage).
After consulting, all the types of the effect were considered as the effect of an incident.
Table \ref{table:ce-entities} show CE's definition and the examples of cause and effect types.

\begin{table}[!h]
\centering
\tiny
\caption{The guideline of CE annotation.}
\label{table:ce-entities}
\begin{tabular}{p{1.2cm}p{5.5cm}} \hline
CE types & Definition \\ \hline
Event\_Leak (EL) & Tag sentences in which gas leakage can be directly confirmed. However, automatic detection by equipment is not included due to the possibility of malfunction. Human detection is included. The definition of gas follows the NER Product. Example: Hydrogen and aniline leakage. \\ \hline

Damage\_Property (DP) & Tag sentences that confirm physical damage to equipment or facilities caused by Event\_Leak and Event\_others. Physical damage includes burst pipes, destruction of heat exchangers, etc. Example: Container ruptures. \\ \hline

Damage\_Human (DH) & Tag sentences that confirm Human casualties caused by Event\_Leak and Event\_others and Damage\_Property. Human casualties include deaths, injuries, and physical illnesses. Example: One employee injured left thigh and left ear. \\ \hline

Event\_others (EO) & Tag sentences containing accident events other than gas leakage. For example, explosions, fires, etc. Example: It is estimated that hydrogen, which has a low ignition energy, was ignited by static electricity. \\ \hline

Cause & Tag sentences that confirm the event causing Event\_Leak and Event\_others. Target not only direct causes but also indirect causes (e.g., Cause's Cause).
In case of ignition or explosion, the three elements of combustion (combustibles, oxygen, and heat) shall be noted as a cause. Example: As a result of reduced tightening torque in some of the flange sections cooled by hydrogen \\

\hline
\end{tabular}
\end{table}

%After finishing the first step with 100 random samples, the agreement score was 0.715 of F-1 (std: 0.044) for the exact match and 0.889 of F-1 (std: 0.009) for token match.
%For the second step, the guideline was applied to annotate the remaining 870 samples divided in three parts, corresponding to the three annotators.
%The agreement score was slightly lower, in which the F-1 was 0.687 (std: 0.024) for the exact match and 0.824 (std: 0.007) for the token match.
%It shows that the exact match is more challenging than the token match for the span level.

The annotation of causes and effects is on the span level.
%and follows the annotation process mentioned in Section \ref{subsec:data-annotation}.
The annotation was done in two steps (follows Section \ref{subsec:data-annotation}), in which the first step was conducted in several rounds to create the annotation guideline to annotate the whole CE dataset.
For annotation, the definition of causes and effects in Table \ref{table:ce-entities} and the rules in Appendix \ref{apx:ce-rules} were used.
After creating the guideline with a high agreement score, the guideline was adopted to annotate the remaining 870 samples.
Fleiss' Kappa score of randomly cross-checking 100 other samples was 0.764.
Table \ref{table:ce-dataset} shows the information of the CE dataset.
467 samples have cause-effect pairs. Others only contain causes or effects.
A sample of cause-effect annotation is shown in Figure \ref{fig:ce-sample}, Appendix \ref{apx:ce-rules}.

\begin{table}[!h]
\centering
\setlength{\tabcolsep}{5.5pt}
\caption{Statistics of the CE dataset.}
\label{table:ce-dataset}
\begin{tabular}{lccc} \hline
Information type & Train & Test & All\\
\hline
Cause & 1,073  & 396 & 1,469 \\
Effect & 1,063  & 400 & 1,463 \\ \hline
\#samples in total & 700 &  270 & 970 \\
\#samples with CE pairs & 467 &  --- & --- \\ \hline
\end{tabular}\vspace{-0.2cm}
\end{table}

\paragraph{IR annotation}\label{para:IR}
The objective of the IR annotation task is to realize a use case where users can query incident descriptions to retrieve relevant past incidents.
%specifically within the context of high-pressure gas scenarios.
We found that the annotation of IR is challenging to measure the similarity of incidents by using single aspects, e.g., the description of incidents. Therefore, instead of directly assigning a relevance score to predefined levels like "Not Relevant," "Relevant," and "Highly Relevant," we first identified a set of key attributes to each incident report and then evaluated relevance on an attribute-by-attribute basis. The attributes allow us to reflect the nature of similarity among incidents.

%dopted a more nuanced strategy. We started by identifying a set of 

We collaborated with domain experts to identify crucial attributes for determining how similar incident reports are. These specific attributes are shown in Table \ref{table:ir-attr}. 
Each incident description was annotated by assigning a relevant label to every identified attribute. The relevance score among incidents was measured by the degree of overlap in their labels. This strategy offers two advantages: (i) it provides a framework for a numeric evaluation of the relevance among incidents and (ii) it allows the flexible generation of relevance scores.

%The latter is accomplished by assigning weights to the shared attributes, reflecting their varying degrees of importance from multiple critical standpoints.

% This labeling enabled us to quantify the correlation among incident descriptions through the overlap degree of attribute labels, and thereby calculate the correlation score between each incident case and all other incident cases at once. The added benefits of this method include fostering computational flexibility in deriving relevance scores by assigning weights to the intersecting attributes of perspectives deemed vital for correlation.
% Following these definitions, annotators assigned labels to the attributes related to each perspective for a given incident case.

\newcommand{\multirowcell}[2][c]{%
\begin{tabular}[#1]{@{}l@{}}#2\end{tabular}}
\begin{table}[!h]
    \centering
    \tiny
    \caption{The definitions of attributes and their labels.}
    \label{table:ir-attr}
    \begin{tabular}{p{1.5cm}p{3.6cm}p{1.2cm}}
        \toprule
        Attribute & Label & \#samples  \\
        \midrule
        \multirow{4}{*}{\multirowcell{Type of high \\pressure gas}}         & (a) Flammable or Flame Retardant Gas      & 911 \\
                                                                            & (b) Toxic Gas                             & 78 \\
                                                                            & (c) Satisfies a \& b                      & 563 \\
                                                                            & (d) Not applicable                        & 607 \\
        \midrule
        \multirow{4}{*}{\multirowcell{Cause of \\incident}}                 & (a) Equipment Factor                      & 940\\
                                                                            & (b) Human Factor                          & 598 \\
                                                                            & (c) External Factor                       & 67 \\
                                                                            & (d) Other Factor                          & 554 \\
        \midrule
        \multirow{6}{*}{\multirowcell{Incident result}}                     & (a) Leakage                               & 1510 \\
                                                                            & (b) Fires and explosions                  & 337 \\
                                                                            & (c) a \& property damage                  & 24 \\
                                                                            & (d) a \& human casualties                & 88 \\
                                                                            & (e) b \& property damage                & 47 \\
                                                                            & (f) b \& human casualties                  & 78 \\
                                                                            & (g) Property damage \& human casualties  & 30 \\
                                                                            & (h) Others  & 45 \\
        \midrule
        \multirow{3}{*}{\multirowcell{Time span from\\ cause to effect}}    & (a) Sudden                                & 364 \\
                                                                            & (b) Long term                             & 1238 \\
                                                                            & (c) Unknown                               & 557 \\
        \midrule
        \multirow{3}{*}{\multirowcell{Operational status \\of equipment}}   & (a) Steady-state operation                & 900 \\
                                                                            & (b) Non-steady state operation  & 344 \\
                                                                            & (c) During maintenance & 409 \\
                                                                            & (d) Other situations                      & 506 \\           
        \bottomrule
    \end{tabular}%\vspace{-0.1cm}
\end{table}

In this study, we analyzed 2,159 high-pressure gas incident reports, detailed in Section \ref{subsec:data-collection}. We employed a straightforward approach where each attribute's label overlap was scored as 1, and no overlap received a score of 0. When labels were jointly attributed through the use of `and'—for instance, label (c) in the \textit{Incident Result type}—the overlap score was increased to 1.5. The final relevance score was computed by summing these individual overlap scores. Sample incident descriptions and their corresponding relevance scores are presented in Figures \ref{fig:ir-sample} and \ref{fig:ir-sample-score}, respectively. In this schema, the incident description itself is used as a query, and the goal of the retriever model is to identify reports with high relevant scores.
To assess inter-annotator reliability, we evaluated the consensus across 100 incident reports annotated by three individuals. The resulting average Fleiss' Kappa score was 0.541, denoting a moderate level of agreement that is good enough for IR.
%with is good enough for IR.

% \begin{table}[!h]
% \centering
% \caption{Statistics of the NER dataset.}
% \label{tab:ir}
% \begin{tabular}{lccc} \hline
% Entity & Train & Test & All\\
% \hline
% Product & 2,189 &  905 & 3,094 \\
% Chemical & 1,440 &  576 & 2,016 \\
% Storage & 5,881  & 2,251 & 8,132 \\
% Incident & 5,274  & 2,045 & 7,319 \\
% Process & 1,138 &  426 & 1,564 \\
% Test & 1,615 & 572  & 2,187 \\
%  \hline
% \#entities in total & 17,537 &  6,775 & 24,312 \\
% \#sentences in total & 700 &  270 & 970 \\ \hline
% \end{tabular}
% \end{table}
% The created dataset is small but very high quality annotated by domain experts. It costs one million Japanese Yen (around 7,170 USD) in five months.

\subsection{Quantitative Observation}
This section shows the statistics of recent incident databases and corpora. The databases include CVE (Common Vulnerabilities and Exposures) \cite{MITRE}, FAA\footnote{We could not know exactly the number of samples.} (Federal Aviation Administration and National Aeronautics and Space Administration) \cite{Federal}, AIID\footnote{https://incidentdatabase.ai} \cite{Sean-database-AAAI-21}, and EF (explosion and fire) \cite{EPFire}. The corpora contain CFDC (high-level causes of flight delays and cancellations) \cite{miyamoto2022natural} and AIR (aviation incident reports) \cite{jiao2022classification}. We also note that there are quite a lot of other incident databases and corpora but due to space limitation, we could not show them all.

%USDrug \cite{USDrug}

\begin{table}[!h]
\centering
\setlength{\tabcolsep}{3.2pt}
\small
\caption{Statistics of incident databases and corpora. Manufac is Manufacturing and Lang is language.}\label{tab:incident-corpora}
\begin{tabular}{lccccc} \hline
Name & Samples & Label & Problem & Domain & Lang\\ \hline
CVE & 141076 & No & IR & Security & EN \\
FAA & --- & No & IR & Aviation & EN \\
%USDrug & --- & --- & -- & Drug & EN \\
AIID & 2842 & No & IR & Mix & EN \\
EF & 6430 & No & IR & Fire & JA \\
CFDC & 4195 & No & Clustering & Aviation & EN \\
AIR & 1775 & Yes & Classification & Aviation & CN \\
\textbf{Ours} & \textbf{970} & \textbf{Yes} & \textbf{NER, CE, IR} & \textbf{Manufac} & \textbf{JA} \\
\hline
\end{tabular}
\end{table}

As observed, incident databases are usually designed for IR in diverse sectors without clear labels. Annotated corpora, e.g., AIR, are created for target problems with a smaller number of samples. Our dataset contains a quite small number of samples. However, it has the human annotation of three NLP tasks which are beneficial for the analysis of incident reports.
In addition, a small number of samples is still helpful in business scenarios for training AI models by using transfer learning \cite{devlin2019bert,nguyen2020transfer,nguyen2023gain}.

%United States Food and Drug Administration

\section{NLP Tasks and Methodology}
Once the dataset has been created, NLP tasks were designed to establish the baselines of each task.

\subsection{Nested Named Entity Recognition}
The NER task was formulated as a sequence labeling problem \cite{ju2018neural,rojas2022simple,zhang2022optimizing,yan2022embarrassingly}.
Strong nested NER models were selected as follows.

\paragraph{Layered nested NER}
This model stacks flat NER layers for nested NER \cite{ju2018neural}. Each flat layer composes of a BiLSTM layer to capture the sequential context representation of an input sequence and a cascaded CRF layer for labeling.
%The output of the BiLSTM in the current flat NER layer is merged to build new representation of the next flat NER layer. The model combines character and token embeddings for representation.

\paragraph{Multiple BiLSTM-CRF}
This model uses multiple flat BiLSTM-CRF, one for each entity type \cite{rojas2022simple}. The input layer combines character embeddings and token representation from Flair \cite{akbik2018contextual} and BERT \cite{devlin2019bert}. The combined representation is fed into BiLSTM layers to obtain long-contextual information. Sequence labeling is done with CRF.

\paragraph{BINDER}
is an optimized bi-encoder model for NER by using contrastive learning \cite{zhang2022optimizing}. It formulates the NER task as a representation learning problem that maximizes the similarity between an entity mention and its type.
%the vector representations of

\paragraph{CNN-Nested-NER}
It is a simple but effective model for nested NER \cite{yan2022embarrassingly}. It uses BERT \cite{devlin2019bert} for mapping input sequences into contextual vectors. The spatial relations among tokens are modeled by an additional CNN layer for prediction with a sigmoid layer.

\paragraph{Preliminary results}
Table \ref{tab:ner-results} reports the performance of the baseline in terms of micro and macro F-scores.
\begin{table}[!h]
\centering
\setlength{\tabcolsep}{3pt}
\caption{Performance of NER models.} \label{tab:ner-results}
\begin{tabular}{lcc} \hline
Model &	 Micro F-1 & Macro F-1 \\ \hline
Nested NER & 78.87 & 75.63 \\
Multiple BiLSTM-CRF     & 83.62 & 79.67 \\
BINDER & 86.96 & 84.02 \\
CNN-Nested-NER & \textbf{87.53} & \textbf{84.54} \\ \hline

\end{tabular}
\end{table}
It shows that the CNN-Nested-NER model is the best for recognizing nested incident entities. A possible reason comes from the use of partial relations among entities and contextual representation from BERT. The BINDER model follows with tiny margins. It shows the contribution of contrastive learning. Two nested NER models based on multiple layers do not show the efficiency. It suggests to improve representation learning.
We did not use LLMs for NER due to nested entities.

\subsection{Cause-Effect Extraction}
The CE extraction task was formulated as a span extraction problem \cite{devlin2019bert,nguyen2023emotion}.
As shown in Table \ref{table:ce-entities}, each sample may contain one or more spans annotated as \textbf{Cause} or other types.
For simplicity, EL, DP, DH, and EO spans were merged as \textbf{Effect} and cause spans were kept identically.
For span-based extraction models, the question is \textit{``cause"} or \textit{``effect"} and the context is an incident report.
This is because the definition of complete questions does not guarantee the semantic relationship between the questions and context documents \cite{mengge2020coarse}.

%\textbf{Event\_Leak (EL)}, \textbf{Event\_Others (EO)}, \textbf{Damage\_Property (DP)}, \textbf{Damage\_Human (DH)}. For simplicity, EL, EO, DP, DH spans were merged as \textbf{Effect}.
%because causes and effects were annotated on the span level.

%It is possible to use pre-defined questions for QA \cite{mengge2020coarse}, however, the definition of questions is time-consuming, needs domain knowledge, and does not guarantee the semantic relationship between the questions and context documents.
%Instead, we use two short questions \textit{``cause}" and \textit{``effect}" as an implicit indicator that provides additional information for BERT-QA.
%Several strong span-based extraction models were implemented for evaluation as follows.

%A sample of high-gas incidents usually contain some causal information that tell us more clearly about what causes the incident and what is the consequence. By applying cause-effect extraction techniques to high-gas incidents, we aim to estimate what are main cause and effect patterns and how well causes and effects are related to each other.

\paragraph{BERT-QA}
We followed BERT-QA \cite{devlin2019bert} to extract cause and effect spans.
The question and the context were concatenated before being encoded by BERT. The contextual representations of tokens were put into a feed-forward network followed by a softmax layer. Each candidate span for the answer was extracted based on \textit{start/end} probabilities predicted by the model.
%The maximum scoring span is used as the prediction.
%The training objective is the loglikelihood of the correct start and end positions.
% \begin{figure}[!h]
% \centering
% \epsfig{file=figures/bert-qa.eps, height=2.0in}
% \setlength{\abovecaptionskip}{5pt}
% \caption{BERT-based extractive Question Answering}
% \label{fig:bert-qa}
% \end{figure}

%Figure \ref{fig:bert-qa} reproduced from \cite{devlin2018bert} shows how BERT is applied to the extractive QA task.
%Tokens of question $q = q_1,..,q_n$ and context $C=c_1,..,c_m$ are concatenated before being encoded by BERT. The contextual representations of tokens $T_i$ are put into a feed-forward layer followed by a softmax. Each candidate span for the answer is scored as the product of \textit{start/end} probabilities. The maximum scoring span is used as the prediction. The training objective is the loglikelihood of the correct start and end positions.

%Because no emotion/cause information is provided beforehand, we have to detect them first with generic questions. It is possible to use pre-defined questions for extraction \cite{mengge2020coarse}, however, we argue that the definition of questions is time-consuming, needs domain knowledge, and does not guarantee the semantic relationship between the questions and context documents. Instead, we use two short questions "\texttt{cause}" and "\texttt{effect}" as an implicit indicator that provides additional information for BERT-QA.

\paragraph{FastQA}
Apart from BERT-QA, we also tested FastQA \cite{son2022jointly}. 
While BERT-QA extracts each cause or effect span independently, FastQA extracts cause and effect simultaneously as a pair. By embedding both \textit{``cause"} and \textit{``effect"} questions in a separate module, FastQA allows the model to encode cause and effect at the same time, which halves the complexity of the encoding. 

\paragraph{Guided-QA}
Guided-QA \cite{nguyen2023emotion} is an extension of BERT-QA that implicitly models the relationship between causes and effects in a sequence manner.
It receives a cause question (\textit{``cause}") and predicts the corresponding cause span. Then the predicted cause span is used as a question for effect extraction. Compared to BERT-QA, Guided-QA takes into account an implicit relationship from effect for cause prediction.

%, i.e., if causes and effects have strong correlation, the detection of causes may help the detection of effects and vice versa.

%FastQA and Guided-QA have the same limitation: each sample must contain at least one cause and one effect. Out of 700 train samples, there are only 467 samples that satisfy this requirement.

\paragraph{LLMs}
We tested ChatGPT\footnote{https://platform.openai.com/playground} and Vicuna-13b-4bit\footnote{https://huggingface.co/elinas/vicuna-13b-4bit} to assess the capability of LLMs for CE in two settings: zero-shot and 1-shot.
For zero-shot experiments, an incident report was appended to a pre-defined prompt such as \textit{"Find the cause and effect in the following incident. Outputs should be in Japanese."} and feed them directly to LLMs. For 1-shot experiments, we used the completion format <instruction><example><input> as follows. 

\texttt{Outputs should be in Japanese.\\
Text: <example incident>\\
Cause/effect: <example cause>\\
Text: <target incident>\\
Cause/effect:}

%-13b-4bit

%\paragraph{Evaluation metrics}
%Because a sample of high-gas incident may contain zero, one or several causes and effects, we count the number of valid causes when the test sample contains at least one cause and the predicted cause is not empty. The same counting rule applies to effects.

%Given ground-truth spans for cause and a predicted span of cause, we compute F1 score between them using the standard BERT-QA metric \cite{devlin2018bert}.

\paragraph{Preliminary results}
Table \ref{tab:ce-results} reports the results of CE models in terms of SQUAD F-1 (token match) \cite{devlin2019bert}.
With full 700 training samples, BERT-QA is competitive followed by ChatGPT 1-shot. It is understandable that BERT-QA was trained with 700 samples and easy for domain adaptation. ChatGPT and Vicuna may need more samples for working well with this CE task.
\begin{table}[!h]
\centering
\caption{Performance of cause-effect extraction models.} \label{tab:ce-results}
\begin{tabular}{lccc}
\hline
Model &	Cause &  Effect & Average \\ \hline
\multicolumn{4}{l}{The train set of 700 samples}\\ \hline
BERT-QA        & 65.90 	  & 77.81 & \textbf{71.85} \\
ChatGPT 1-shot & 75.48    & 49.80 & 62.64 \\
ChatGPT 0-shot & 55.48 	  & 37.25 & 46.36 \\
Vicuna 1-shot  & 32.00    & 31.55 & 31.77 \\
Vicuna 0-shot  & 19.44    & 28.16 & 23.80 \\ \hline
\multicolumn{4}{l}{The train set of 467 samples}\\ \hline
BERT-QA        & \textbf{76.34} & \textbf{80.98} & \textbf{78.66} \\
FastQA         &	71.25  & 79.39 & 75.32 \\
Guided-QA      & 	75.81  &	72.72 & 74.26 \\
\hline
\end{tabular}%\vspace{-0.2cm}
\end{table}

As mentioned in Table \ref{table:ce-dataset}, 467 samples have cause-effect pairs. We did another experiment on this refined set. The F-scores show that span-based extraction models obtain improvement compared to the models trained on the original set. It shows that with more refined training samples, a simple BERT-QA model can achieve promising results. Note that FastQA and Guided-QA can only work with samples that include cause-effect pairs.

% \begin{table*}[!t]
% \centering
% \caption{Performance of QA-based baselines and LLM baselines} \label{tab:cause-effect-scores}
% \begin{tabular}{|c|r|r|r|r|}
% \hline
% Model & Cause (\#valid) &	Cause (F1) &	Effect (\#valid) 	& Effect (F1) \\
% \hline
% BERT-QA             & 191/270 	    & 0.659 	  & 259/270 	& 0.778 \\
% ChatGPT 1-shot      & 192/270       & 0.755         &	261/270 & 0.498 \\
% ChatGPT 0-shot      & 190/270       & 0.555 	   & 227/270 	& 0.372 \\
% Vicuna-13b-4bit 1-shot     & 79/270        & 0.320     & 147/270       & 0.315 \\
% Vicuna-13b-4bit 0-shot     &  8/270        & 0.194      & 41/270        & 0.282 \\
% \hline
% \multicolumn{5}{|l|}{train set of 467 samples}\\
% \hline
% BERT-QA             & 192/270       &	\textbf{0.763} & 	261/270 & 	\textbf{0.810} \\
% FastQA              & 191/270       &	0.713         & 261/270 	& 0.794 \\
% Guided-QA           & 192/270       & 	0.758            &	261/270   &	0.727 \\
% \hline
% \end{tabular}%\vspace{-0.2cm}
% \end{table*}

\subsection{Information Retrieval}
The IR task was formulated as a dense text retriever problem using bi-encoder \cite{zhao2022dense}. 
A deep neural network was used to convert incident reports and queries into dense vectors with their nearest neighbors searched in the database.
We conduct the evaluation on the annotated IR dataset with several baselines as follows.

%To search for similar historical events in the database, we first employs dense vectors (embeddings) from a deep neural network to represent the context. Then, a query is encoded with the same model to obtain the query vector, which is followed by a nearest neighbors search in the database. We conduct the evaluation on the annotated IR data with several baselines as follows.

\paragraph{BERT-based bi-encoder (public)}
Bi-encoders are highly efficient retrieval models based on pretrained transformer backbones (e.g., BERT). We utilized the popular sentence-BERT multilingual model\footnote{https://huggingface.co/sentence-transformers/distiluse-base-multilingual-cased-v2} \cite{nilsbert} as the main baseline for our dense retrieval task.

\paragraph{BERT-based bi-encoder (finetuned)}
To better adapt the model to challenging technical terms and jargon in the incident reports, we further fine-tuned the aforementioned base encoder by using the unsupervised constrastive learning objective \cite{gaosimcse} on the collection of the incident corpus in Section \ref{para:IR}. Detail of the fine-tuning process can be found in Appendix \ref{apx:ir}.

\paragraph{Commercial embedding model (OpenAI)}
We also evaluated the recent commercial solution from OpenAI with the model name \texttt{text-embedding-ada-002}.\footnote{https://platform.openai.com/docs/guides/embeddings/what-are-embeddings} The model is available in form of an API, which we can use to create the embedding vector for a given document.

\paragraph{Preliminary results}
Table \ref{tab:ir-results} presents the results of all IR models with \texttt{nDCG@k}, \texttt{mAP@k} and \texttt{Recall@k} as evaluation metrics with \texttt{k=20}. The evaluation dataset is described in Section \ref{subsec:data-annotation}.
\begin{table}[!h]
\centering
\setlength{\tabcolsep}{5pt}
\caption{Performance of information retrieval models.} \label{tab:ir-results}
\begin{tabular}{lccc}
\hline
Model &	nDCG@k &  mAP@k & R@k \\ \hline
BERT-public        & 45.27 	  & 15.91 & 30.90 \\
BERT-finetuned & \textbf{56.11}    & 21.72 & \textbf{42.51} \\
OpenAI-emb & 54.48 	  & \textbf{22.25} & 38.32 \\
\hline
\end{tabular}%\vspace{-0.2cm}
\end{table}

We can observe from the table that the fine-tuned BERT encoder produces significantly better performance than the default base model and achieves the best score on \texttt{Recall@k} and \texttt{nDCG@k}. The OpenAI embedding model closely follows the fine-tuned model, albeit not directly trained on similar data domains before. This shows the performance of the proprietary model from OpenAI is quite transferable and robust across different domains.

\subsection{Output Observation}
%We observe the output of NER, CE, and IR. Detailed outputs are shown in Appendix \ref{app:output-observation}.

\paragraph{Nested NER}
Figure \ref{fig:ner_ce_output}(a) shows the example of a success case, where the model correctly detects all spans of entities, including the nested one between Product and Test. Further observation shows that for the same entities that are close together, the model tends to incorrectly recognize these entities separately, as shown in figure \ref{fig:ner_ce_output}(b). This type of error is more common for entities such as Incident and Process due to their complex nature and their lengths. For entities such as Chemical and Product, the common problem is misclassification of entities or incorrect recognition of other noun spans. The observation was done by using CNN-Nested-NER.

\paragraph{Cause-Effect Extraction}
As we observe the data, cause and effect spans usually appear in phrases that indicate incidents such as leak, insufficient tightening. Because causes and effects share such common patterns, it is harder for our models to make correct predictions. Figures \ref{fig:ner_ce_output}(a) and (b) show an example of correct effect prediction and an example of incorrect cause prediction of BERT-QA finetuned on 467 samples (the best model).

\paragraph{Information Retrieval}
We analyze the success and failure cases of IR model BERT-finetuned. Most success retrieval cases such as Figure \ref{fig:ir-success}, with query document at the top most and followed retrieval results, typically mention several common subjects such as flame, gas leak, and substance name (ethylene). However, there are still a lot of failure cases of the model regarding the understanding of substance properties (toxic, flammable, etc) when retrieving similar cases containing different substances, and understanding the effect (e.g: leakage vs explosion) of the incident (Figure \ref{fig:ir-failure}).

\section{Conclusion}
This paper introduces a new Japanese dataset for safety prevention by using AI models. The high-quality dataset is annotated by domain experts for NER, CE, and IR tasks.
The dataset contributes to IncidentAI in two important points.
First, it composes the three NLP tasks in a corpus that facilitates the development of AI pipelines for safety prevention in a low-resource language.
Second, it benchmarks the results of the three tasks which are beneficial for the next studies of analyzing incident reports.
Future work will adapt the dataset to create AI pipelines for preventing failures of IncidentAI.

\section*{Limitations}
Although the newly created dataset of incidents is a very high-quality corpus that is composed of three NLP tasks: NER, cause-effect extraction (CE), and IR, the size of the dataset is quite small with 970 annotated samples for NER and CE. The number of annotated samples for IR is also small with 2,159 samples.
While collecting raw data is quite easy, data annotation is time-consuming and labor-expensive with the involvement of domain experts. It explains the size of our dataset is quite limited.
So, it requires more effort for data augmentation when using the dataset in some cases. For example, LLMs need thousands annotated samples for fine-tuning.
In addition, the dataset is in Japanese. On the one hand, it facilitates the introduction of AI models for IncidentAI in a low-resource language. However, the dataset requires translation to more popular languages, e.g., English for wider use.

For evaluation, some models are quite straightforward because the purpose is to provide preliminary results of the dataset. We believe the performance of the three tasks can be still improved with stronger models, especially in the case of cause-effect extraction with BERT-QA and LLMs.

\section*{Ethics Statement}
The dataset and models experimented in this work have no unethical applications or risky broader impacts.
The dataset was crawled from publicly available reports of high-pressure gas incidents published in 2022 by the High Pressure Gas Safety Institute of Japan. Raw data contains information such as descriptions of incidents at high-pressure gas plants, types of incidents, dates of incidents, industries, ignition sources, etc. It does not include any confidential or personal information of workers or companies. Three annotators are domain experts who have at least six years of experience in the high-pressure gas incident domain. They knew the purpose of data creation and agreed to join the annotation process with their responsibilities. Their personal information is kept for data publication.

The models used for evaluation can be publicly accessed with GitHub links. There is no bias for the re-implementation that can affect the final results.

\bibliography{incident_ai}

\begin{thebibliography}{38}
\expandafter\ifx\csname natexlab\endcsname\relax\def\natexlab#1{#1}\fi

\bibitem[{Administration(2023)}]{Federal}
Federal~Aviation Administration. 2023.
\newblock Accident and incident data.
\newblock \url{https://www.faa.gov/data_research/accident_incident}.
\newblock Accessed: 2023-07-17.

\bibitem[{Akbik et~al.(2018)Akbik, Blythe, and Vollgraf}]{akbik2018contextual}
Alan Akbik, Duncan Blythe, and Roland Vollgraf. 2018.
\newblock Contextual string embeddings for sequence labeling.
\newblock In \emph{Proceedings of the 27th international conference on
  computational linguistics}, pages 1638--1649.

\bibitem[{Andrade and Walsh(2023)}]{andrade2023safeaerobert}
Sequoia~R Andrade and Hannah~S Walsh. 2023.
\newblock Safeaerobert: Towards a safety-informed aerospace-specific language
  model.
\newblock In \emph{AIAA AVIATION 2023 Forum}, page 3437.

\bibitem[{Chandra et~al.(2023)Chandra, Jing, Bendarkar, Sawant, Elias, Kirby,
  and Mavris}]{chandra2023aviation}
Chetan Chandra, Xiao Jing, Mayank~V Bendarkar, Kshitij Sawant, Lidya Elias,
  Michelle Kirby, and Dimitri~N Mavris. 2023.
\newblock Aviation-bert: A preliminary aviation-specific natural language
  model.
\newblock In \emph{AIAA AVIATION 2023 Forum}, page 3436.

\bibitem[{Corporation(2023)}]{MITRE}
The~MITRE Corporation. 2023.
\newblock Cve - common vulnerabilities and exposures.
\newblock \url{https://cve.mitre.org/cve/search_cve_list.html}.
\newblock Accessed: 2023-07-17.

\bibitem[{Devlin et~al.(2019)Devlin, Ming-Wei, Lee, and
  Toutanova}]{devlin2019bert}
Jacob Devlin, Changm Ming-Wei, Kenton Lee, and Kristina Toutanova. 2019.
\newblock Bert: Pre-training of deep bidirectional transformers for language
  understanding.
\newblock In \emph{Proceedings of NAACL-HLT}, pages 4171--4186.

\bibitem[{Durso et~al.(2022)Durso, Raunak, Kuhn, and
  Kacker}]{durso2022analyzing}
Francis Durso, MS~Raunak, Rick Kuhn, and Raghu Kacker. 2022.
\newblock Analyzing failures in artificial intelligent learning systems
  (fails).
\newblock In \emph{2022 IEEE 29th Annual Software Technology Conference (STC)},
  pages 7--8. IEEE.

\bibitem[{Food and Administration(2023)}]{USDrug}
United~States Food and Drug Administration. 2023.
\newblock Accident and incident data.
\newblock
  https://www.fda.gov/drugs/questions-and-answers-fdas-adverse-event-reporting-system-faers/fda-adverse-event-reporting-system-faers-public-dashboard.
\newblock Accessed: 2023-07-17.

\bibitem[{Gao et~al.(2021)Gao, Yao, and Chen}]{gaosimcse}
Tianyu Gao, Xingcheng Yao, and Danqi Chen. 2021.
\newblock \href {http://arxiv.org/abs/2104.08821} {Simcse: Simple contrastive
  learning of sentence embeddings}.
\newblock volume abs/2104.08821.

\bibitem[{Hong et~al.(2021)Hong, Fourney, DeBellis, and
  Amershi}]{hong2021planning}
Matthew~K Hong, Adam Fourney, Derek DeBellis, and Saleema Amershi. 2021.
\newblock Planning for natural language failures with the ai playbook.
\newblock In \emph{Proceedings of the 2021 CHI Conference on Human Factors in
  Computing Systems}, pages 1--11.

\bibitem[{Jiao et~al.(2022)Jiao, Dong, Han, and Sun}]{jiao2022classification}
Yang Jiao, Jintao Dong, Jingru Han, and Huabo Sun. 2022.
\newblock Classification and causes identification of chinese civil aviation
  incident reports.
\newblock \emph{Applied Sciences}, 12(21):10765.

\bibitem[{Jing et~al.(2023)Jing, Chennakesavan, Chandra, Bendarkar, Kirby, and
  Mavris}]{jing2023bert}
Xiao Jing, Akul Chennakesavan, Chetan Chandra, Mayank~V Bendarkar, Michelle
  Kirby, and Dimitri~N Mavris. 2023.
\newblock Bert for aviation text classification.
\newblock In \emph{AIAA AVIATION 2023 Forum}, page 3438.

\bibitem[{Ju et~al.(2018)Ju, Miwa, and Ananiadou}]{ju2018neural}
Meizhi Ju, Makoto Miwa, and Sophia Ananiadou. 2018.
\newblock A neural layered model for nested named entity recognition.
\newblock In \emph{Proceedings of the 2018 Conference of the North American
  Chapter of the Association for Computational Linguistics: Human Language
  Technologies, Volume 1 (Long Papers)}, pages 1446--1459.

\bibitem[{Macrae(2022)}]{macrae2022learning}
Carl Macrae. 2022.
\newblock Learning from the failure of autonomous and intelligent systems:
  Accidents, safety, and sociotechnical sources of risk.
\newblock \emph{Risk analysis}, 42(9):1999--2025.

\bibitem[{McGregor(2021)}]{Sean-database-AAAI-21}
Sean McGregor. 2021.
\newblock Preventing repeated real world ai failures by cataloging incidents:
  The ai incident database.
\newblock In \emph{Proceedings of the AAAI Conference on Artificial
  Intelligence, vol. 35, no. 17, pp. 15458-15463}.

\bibitem[{McGregor et~al.(2022)McGregor, Paeth, and
  Lam}]{McGregor-indexing-AI-risks-22}
Sean McGregor, Kevin Paeth, and Khoa Lam. 2022.
\newblock Indexing ai risks with incidents, issues, and variants.
\newblock In \emph{arXiv preprint arXiv:2211.10384}.

\bibitem[{Mengge et~al.(2020)Mengge, Yu, Zhang, Liu, Zhang, and
  Wang}]{mengge2020coarse}
Xue Mengge, Bowen Yu, Zhenyu Zhang, Tingwen Liu, Yue Zhang, and Bin Wang. 2020.
\newblock Coarse-to-fine pre-training for named entity recognition.
\newblock In \emph{Proceedings of the 2020 Conference on Empirical Methods in
  Natural Language Processing (EMNLP)}, pages 6345--6354.

\bibitem[{Miyamoto et~al.(2022)Miyamoto, Bendarkar, and
  Mavris}]{miyamoto2022natural}
Ayaka Miyamoto, Mayank~V Bendarkar, and Dimitri~N Mavris. 2022.
\newblock Natural language processing of aviation safety reports to identify
  inefficient operational patterns.
\newblock \emph{Aerospace}, 9(8):450.

\bibitem[{Mokhatab et~al.(2018)Mokhatab, Poe, and Mak}]{mokhatab2018handbook}
Saeid Mokhatab, William~A Poe, and John~Y Mak. 2018.
\newblock \emph{Handbook of natural gas transmission and processing: principles
  and practices}.
\newblock Gulf professional publishing.

\bibitem[{Nguyen and Nguyen(2023)}]{nguyen2023emotion}
Huu-Hiep Nguyen and Minh-Tien Nguyen. 2023.
\newblock Emotion-cause pair extraction as question answering.
\newblock In \emph{Proceedings of the 15th International Conference on Agents
  and Artificial Intelligence - Volume 3: ICAART}, pages 988--995.

\bibitem[{Nguyen et~al.(2020)Nguyen, Phan, Linh, Son, Dung, Hirano, and
  Hotta}]{nguyen2020transfer}
Minh-Tien Nguyen, Viet-Anh Phan, Le~Thai Linh, Nguyen~Hong Son, Le~Tien Dung,
  Miku Hirano, and Hajime Hotta. 2020.
\newblock Transfer learning for information extraction with limited data.
\newblock In \emph{Computational Linguistics: 16th International Conference of
  the Pacific Association for Computational Linguistics, PACLING 2019, Hanoi,
  Vietnam, October 11--13, 2019, Revised Selected Papers 16}, pages 469--482.
  Springer.

\bibitem[{Nguyen et~al.(2023)Nguyen, Son et~al.}]{nguyen2023gain}
Minh-Tien Nguyen, Nguyen~Hong Son, et~al. 2023.
\newblock Gain more with less: Extracting information from business documents
  with small data.
\newblock \emph{Expert Systems with Applications}, 215:119274.

\bibitem[{NIOSH(2023)}]{EPFire}
NIOSH. 2023.
\newblock National institute of occupational safety and health, japan.
\newblock
  \url{https://www.jniosh.johas.go.jp/publication/houkoku/houkoku_2020_05.html}.
\newblock Accessed: 2023-10-17.

\bibitem[{Nor et~al.(2022)Nor, Pedapati, Muhammad, and
  Leiva}]{nor2022abnormality}
Ahmad Kamal~Mohd Nor, Srinivasa~Rao Pedapati, Masdi Muhammad, and V{\'\i}ctor
  Leiva. 2022.
\newblock Abnormality detection and failure prediction using explainable
  bayesian deep learning: Methodology and case study with industrial data.
\newblock \emph{Mathematics}, 10(4):554.

\bibitem[{NTSB(2017)}]{national2017collision}
NTSB. 2017.
\newblock Collision between a car operating with automated vehicle control
  systems and a tractor-semitrailer truck.
\newblock \emph{Technical Report No. NTSB/HAR-17/02}.

\bibitem[{Pellegrini et~al.(2019)Pellegrini, DE~GUIDO, Lange
  et~al.}]{pellegrini2019handbook}
Laura~A Pellegrini, Giorgia DE~GUIDO, Stefano Lange, et~al. 2019.
\newblock Handbook of natural gas transmission and processing-principles and
  practices.
\newblock In \emph{Handbook of Natural Gas Transmission and Processing:
  Principles and Practices}, pages 669--739. Elsevier-Gulf Professional
  Publishing.

\bibitem[{Pittaras and
  McGregor(2022)}]{Pittaras-taxonomy-system-cause-analysis-22}
Nikiforos Pittaras and Sean McGregor. 2022.
\newblock A taxonomic system for failure cause analysis of open source ai
  incidents.
\newblock In \emph{arXiv preprint arXiv:2211.07280}.

\bibitem[{Reimers and Gurevych(2019)}]{nilsbert}
Nils Reimers and Iryna Gurevych. 2019.
\newblock \href {http://arxiv.org/abs/1908.10084} {Sentence-bert: Sentence
  embeddings using siamese bert-networks}.
\newblock volume abs/1908.10084.

\bibitem[{Rojas et~al.(2022)Rojas, Bravo-Marquez, and
  Dunstan}]{rojas2022simple}
Mat{\'\i}as Rojas, Felipe Bravo-Marquez, and Jocelyn Dunstan. 2022.
\newblock Simple yet powerful: An overlooked architecture for nested named
  entity recognition.
\newblock In \emph{Proceedings of the 29th International Conference on
  Computational Linguistics}, pages 2108--2117.

\bibitem[{Shrishak(2023)}]{shrishak2023deal}
Kris Shrishak. 2023.
\newblock How to deal with an ai near-miss: Look to the skies.
\newblock \emph{Bulletin of the Atomic Scientists}, 79(3):166--169.

\bibitem[{Son et~al.(2022)Son, Hieu, Nguyen, and Nguyen}]{son2022jointly}
Nguyen~Hong Son, M~Yu Hieu, Tuan-Anh~D Nguyen, and Minh-Tien Nguyen. 2022.
\newblock Jointly learning span extraction and sequence labeling for
  information extraction from business documents.
\newblock In \emph{2022 International Joint Conference on Neural Networks
  (IJCNN)}, pages 1--8. IEEE.

\bibitem[{Tikayat~Ray et~al.(2023)Tikayat~Ray, Cole, Pinon~Fischer, White, and
  Mavris}]{tikayat2023aerobert}
Archana Tikayat~Ray, Bjorn~F Cole, Olivia~J Pinon~Fischer, Ryan~T White, and
  Dimitri~N Mavris. 2023.
\newblock aerobert-classifier: Classification of aerospace requirements using
  bert.
\newblock \emph{Aerospace}, 10(3):279.

\bibitem[{Wolf et~al.(2020)Wolf, Debut, Sanh, Chaumond, Delangue, Moi, Cistac,
  Rault, Louf, Funtowicz et~al.}]{wolf2020transformers}
Thomas Wolf, Lysandre Debut, Victor Sanh, Julien Chaumond, Clement Delangue,
  Anthony Moi, Pierric Cistac, Tim Rault, R{\'e}mi Louf, Morgan Funtowicz,
  et~al. 2020.
\newblock Transformers: State-of-the-art natural language processing.
\newblock In \emph{Proceedings of the 2020 conference on empirical methods in
  natural language processing: system demonstrations}, pages 38--45.

\bibitem[{Yampolskiy(2019)}]{yampolskiy2019predicting}
Roman~V Yampolskiy. 2019.
\newblock Predicting future ai failures from historic examples.
\newblock \emph{foresight}, 21(1):138--152.

\bibitem[{Yan et~al.(2022)Yan, Sun, Li, and Qiu}]{yan2022embarrassingly}
Hang Yan, Yu~Sun, Xiaonan Li, and Xipeng Qiu. 2022.
\newblock An embarrassingly easy but strong baseline for nested named entity
  recognition.
\newblock \emph{arXiv preprint arXiv:2208.04534}.

\bibitem[{Yang and Huang(2023)}]{yang2023natural}
Chuyang Yang and Chenyu Huang. 2023.
\newblock Natural language processing (nlp) in aviation safety: Systematic
  review of research and outlook into the future.
\newblock \emph{Aerospace}, 10(7):600.

\bibitem[{Zhang et~al.(2022)Zhang, Cheng, Gao, and Poon}]{zhang2022optimizing}
Sheng Zhang, Hao Cheng, Jianfeng Gao, and Hoifung Poon. 2022.
\newblock Optimizing bi-encoder for named entity recognition via contrastive
  learning.
\newblock In \emph{The Eleventh International Conference on Learning
  Representations}.

\bibitem[{Zhao et~al.(2022)Zhao, Liu, Ren, and Wen}]{zhao2022dense}
Wayne~Xin Zhao, Jing Liu, Ruiyang Ren, and Ji-Rong Wen. 2022.
\newblock \href {http://arxiv.org/abs/2211.14876} {Dense text retrieval based
  on pretrained language models: A survey}.

\end{thebibliography}
\bibliographystyle{acl_natbib}

\appendix

\section{Appendix}\label{apx:appendix}

\subsection{Annotation rules and examples of NER}\label{apx:ner-rules}
The following rules must be observed when tagging entities from domain experts.
\begin{itemize}
    \item Do not tag words indicating parts such as entrances, exits, and connections if they are at the end of a word. For example, tag "inlet pipe" as "inlet pipe", but for "heat exchanger outlet", only tag "heat exchanger".
    \item For tagging corresponding parts that indicate a range like "4-6", tag the entire "4-6".
    \item If the same word is used in different meanings, tag only the relevant entity.
    \item For words like "XX gas generation equipment," tag both Storage and Products (nested). For example, tag "XX gas generation equipment" as Storage and tag “XX gas”” as Products.
    \item If there is a modifier Process within Products, Chemicals, or Storage, do not tag Process. For example, do not tag "recycle" as Process in "recycle gas."
    \item Do not tag phrases containing particles like "of" in "XX of YY" (Tag only "XX" or "YY" separately).
    \item Tag abbreviations as well. However, do not tag specific abbreviations such as equipment or model numbers.
    \item Do not tag the state of individuals. For example: "Lack of perspective".
    \item Do not tag words within legal names or standards, such as the names of laws or regulations. For example: "High-Pressure Gas Safety Act," do not tag "High-Pressure Gas."
    \item Only tag Pc when it pertains to gas handling. For example: Do not tag "tightening further".
\end{itemize}

\begin{figure*}[!h]
    \centering
    \label{table:ner-example}
    \begin{subfigure}[b]{0.85\textwidth}  % サイズを調整
        \includegraphics[width=\textwidth]{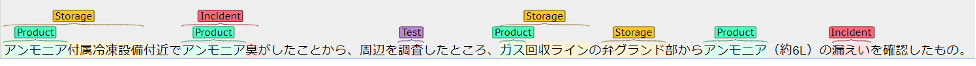}
        \caption{Japanese (original).}
        \label{fig:sub1}
    \end{subfigure}

    \begin{subfigure}[b]{0.85\textwidth}  % サイズを調整
        \includegraphics[width=\textwidth]{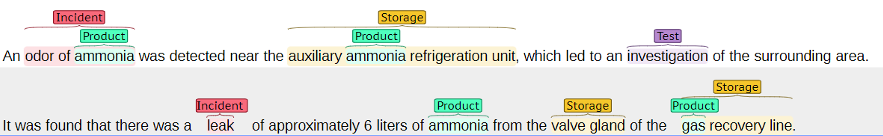}
        \caption{English (translated version).}
        \label{fig:sub2}
    \end{subfigure}

    \caption{An annotated sample of NER.}
    \label{fig:ner-sample}
\end{figure*}

% Figure \ref{fig:ner-sample} shows a translated sample of NER.
% \begin{figure*}[!h]
%     \centering
%     \includegraphics[width=0.9\textwidth]{figures/ner-sample.pdf}
%     \caption{An annotated example of NER translated from an original Japanese sample.}
%     \label{fig:ner-sample}
% \end{figure*}

\subsection{Annotation rules and examples of CE}\label{apx:ce-rules}
The following rules must be strictly followed when tagging CE from domain experts.
\begin{itemize}
    \item Include in one sentence to be tagged: Who, When, Where, What. Also include endings up to verb phrases (e.g. Tag up to "leaked").
    \item Do not include in a sentence to be tagged: (i) punctuation at the end of tagging ", and.", (ii) such as "due to...", and (iii) conjunctions at the beginning of a sentence (e.g. And," "And then," "And then," etc.).
    \item How to separate each tagging: (i) do not separate with "," but separate with ".", (ii) in the case of "broken and leaked," "broken" and "leaked" are two different tags, so separate them.
    \item No nesting.
    \item Do not tag the trigger for accident discovery (unless it is a causal factor in the accident). For example: "I noticed a strange odor,"
\end{itemize}

\begin{figure*}[!h]
    \centering
    \label{table:ce-example}
    \begin{subfigure}[b]{0.85\textwidth}  % サイズを調整
        \includegraphics[width=\textwidth]{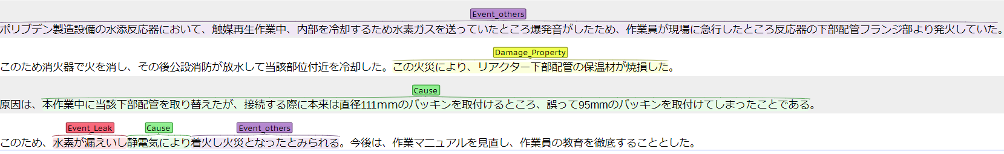}
        \caption{Japanese (original).}
        \label{fig:sub1}
    \end{subfigure}

    \begin{subfigure}[b]{0.85\textwidth}  % サイズを調整
        \includegraphics[width=\textwidth]{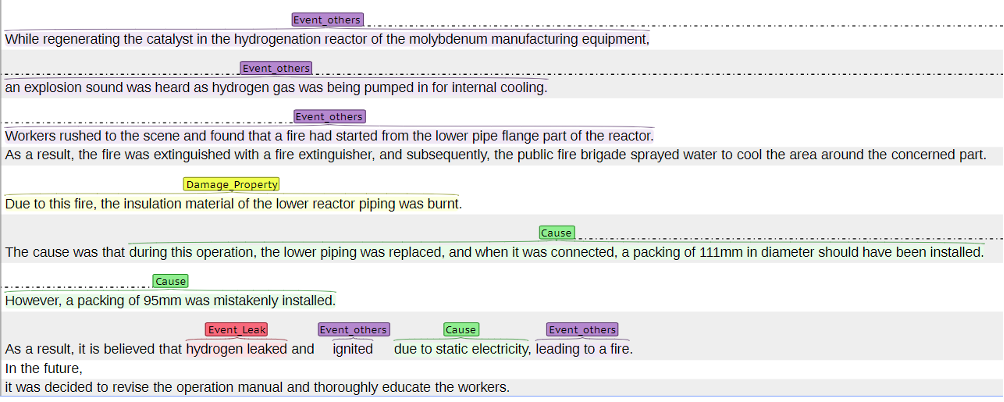}
        \caption{English (translated version).}
        \label{fig:sub2}
    \end{subfigure}

    \caption{An annotated sample of CE.}
    \label{fig:ce-sample}
\end{figure*}

% \begin{table}[!h]
% \centering
% \caption{A translated sample that includes a cause (in \textcolor{red}{red}) and an effect (in \textcolor{green}{green}).}
% \label{table:ce-example}
% \begin{tabular}{|p{2.8in}|} 
% \hline
% %\begin{CJK}{UTF8}{min}
% %2021-320 令和３年８月１０日工場外周パトロール中に，室外機周辺に水漏れがあることを確認する。業者に点検を依頼し，８月１６日空調機を停止させ点検を実施するが，不具合は確認できず。\textcolor{red}{[試運転のため起動させたところ，「ポン」とガスの吹き出し音があり，音の出た付近を確認すると冷媒配管のろう付け部に亀裂を確認したため]}，空調機を停止させる。停止により自動で電磁弁が自動開閉され冷媒回路はブロックされるが，\textcolor{green}{[亀裂のあったブロック範囲のフロンＲ４１０Ａ10.3kgが漏えいしたもの]}。
% %\end{CJK*} 

% 2021-320 August 10, 3rd year of Reiwa. During the patrol around the factory, it was confirmed that there was a water leak around the outdoor unit. We asked a contractor to perform an inspection, and on August 16, we stopped the air conditioner and carried out an inspection, but no problems were found. \textcolor{red}{[When the system was started for a test run, there was a popping sound of gas blowing out, and when the sound was heard, cracks were found in the brazed part of the refrigerant pipe]}, so the air conditioner was stopped. The solenoid valve is automatically opened and closed by stopping and the refrigerant circuit is blocked, but \textcolor{green}{[the cracked block range of CFCs R410A 10.3 kg leaked]}. \\
% \hline
% \end{tabular}
% \end{table}

\subsection{Annotation rules and examples of IR}\label{apx:ir-rules}
\begin{figure*}[!h]
    \centering
    \includegraphics[width=0.9\textwidth]{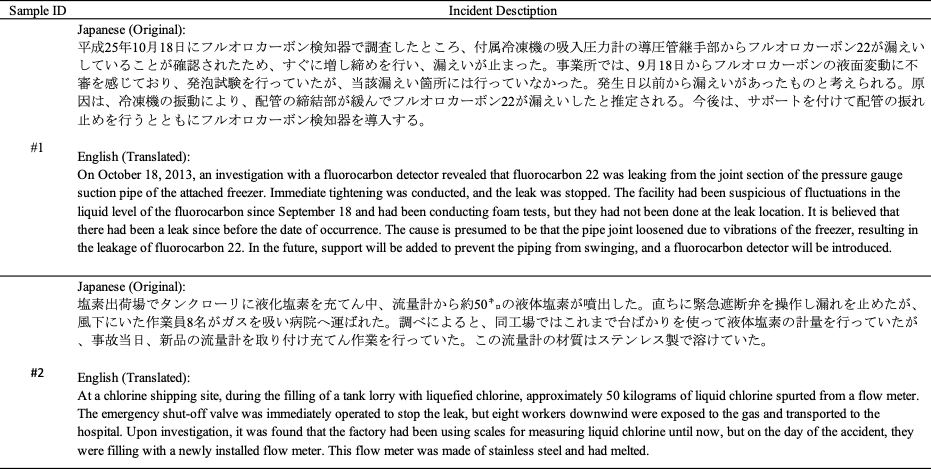}
    \caption{The samples of incident descriptions for IR.}
    \label{fig:ir-sample}
\end{figure*}

Definitions of each attribute are shown in Table \ref{table:ir-definition}. The annotated example of IR is shown in Figure \ref{fig:ir-sample}

\begin{table*}[!h]
    \centering
    \small
    \caption{The guideline of IR annotation.}
    \label{table:ir-definition}
    \begin{tabular}{p{4.5cm}p{10cm}} \hline
    Attribute & Definition \\ \hline
    Type of high pressure gas & The high-pressure gas that caused the reported accident was classified from the perspective of danger in the event of an accident. Cases where the gas could not be identified were included under “d. Not applicable”.
    The definition of flammable gas and toxic gas shall conform to the High Pressure Gas Safety Act in Japan.
     \\ \hline
    
    Cause of incident & The events that caused or triggered the accident were classified. Equipment factors refer to those caused by initial defects in parts built into the equipment. Human factors refer to errors made in operation or judgment by people on site. External factors indicate those caused by events from outside the equipment, such as falling objects.\\ \hline
    
    Incident result & The events that occurred as a result of the accident were classified. Physical and human damage were only considered if they occurred as secondary events, such as gas leaks or fires.
    Property damage: Accidents resulting in damage to equipment or facilities due to fire or explosion.
    Do not include damage to equipment or other items that caused the accident.
    Human casualties: Accidents resulting in health hazards to humans due to leakage, fire, or explosion
     \\ \hline
    
    Time span from cause to effect & The classification was made based on the time from when the cause or trigger of the accident occurred until the accident event took place.
    Sudden: Accidents where the results are caused generally within a few minutes to several tens of minutes from the occurrence of the cause.
     \\ \hline
    
    Operational status of equipment & The classification was made based on the operational status of the equipment at the time of the accident. 
    Non-steady state operation refers to operating conditions that differ from normal operation, such as immediately after the equipment starts running or during test operation
    \\ \hline
\end{tabular}
\end{table*}

\begin{figure*}[!h]
    \centering
    \includegraphics[width=0.9\textwidth]{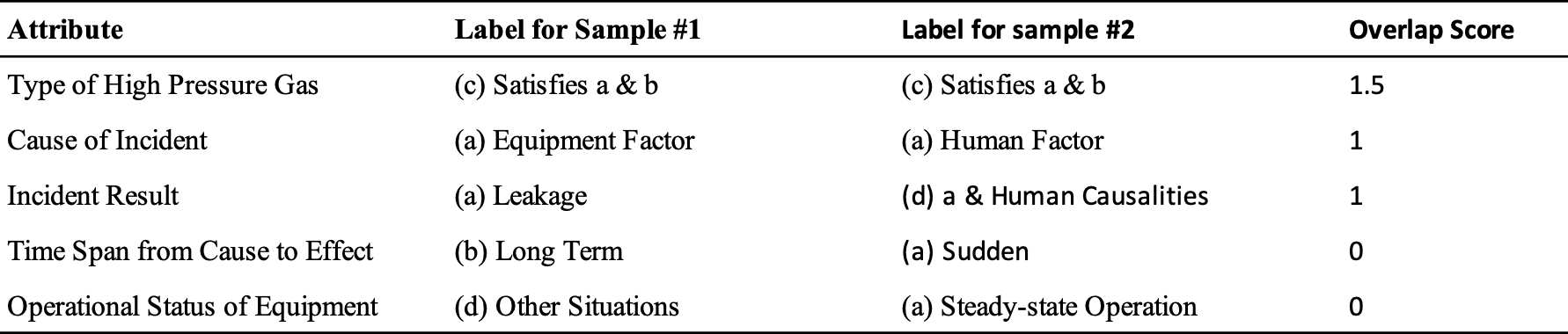}
    \caption{The scoring method for the samples in Figure \ref{fig:ir-sample} is as follows: each single overlap is assigned a score of 1. If a label partially overlaps, as seen with the factor \textit{"a"} under the \textit{"Incident Result"} attribute, it still receives a score of 1. With instances where a label overlaps with two factors, such as with `Type of High Pressure Gas`, the score is 1.5. The final correlation score is the sum of each individual overlap score, totaling \textbf{3.5} in this case.}
    \label{fig:ir-sample-score}
\end{figure*}

% \newcommand{\multirowcell}[2][c]{%
% \begin{tabular}[#1]{@{}l@{}}#2\end{tabular}}
% \begin{table}
%     \centering
%     \tiny
%     \caption{The definitions of Attributes and their labels}
%     \label{table:ir-attr}
%     \begin{tabular}{p{1.7cm}p{4cm}p{0.6cm}}
%         \toprule
%         Attribute & Label &\#samples \\
%         \midrule
%         \multirow{5}{*}{\multirowcell{hoge}}
%             & Type of high pressure gas      & (d) Not applicable \\
%             & Cause of incident              & (a) Equipment factor \\
%             & Incident Result                & (a) Leakage \\
%             & Time span from case to effect  & (b) Long term \\
%             & Operational status of equipment& (a) Steady-state operation \\
%         \bottomrule
%     \end{tabular}
% \end{table}

% \newcommand{\multirowcell}[2][c]{%
% \begin{tabular}[#1]{@{}l@{}}#2\end{tabular}}

\subsection{Implementation}
\subsubsection{NER models}
Except for the neural layered model for nested NER \cite{ju2018neural}, whose word representation is based on the concatenation of character and word embeddings, other models use pre-trained BERT-based encoder, \textit{TurkuNLP/wikibert-base-ja-cased}.\footnote{https://huggingface.co/TurkuNLP/wikibert-base-ja-cased} Other hyperparameters are set as follows.

\paragraph{Layered nested NER}
The number of training epochs is set at 100, with the learning rate and decay rate of $1e-4$ and the batch size of 32. The dimension of word embedding and character embedding are 200 and 25, respectively.

\paragraph{Multiple BiLSTM-CRF}
The number of training epochs is 10 for each entity type.

\paragraph{BINDER}
The number of training epochs is 10, with the learning rate of $3e-5$ and the batch size of 8 due to the heavy model. In addition, the model requires a text description written in Japanese for each entity type, which describes what the entity is and how it is labeled.

\paragraph{CNN-nested-NER}
The number of training epochs is 10, with the learning rate of $3e-5$ and the batch size of 8. The depth of CNN layers is 3, with a dimension of 120 for each.

\subsubsection{Cause-effect extraction models}
The BERT-QA models were implemented using BERT classes provided by Huggingface \cite{wolf2020transformers}. The model was trained in 5 epochs, with the learning rate of $5e-5$, and the batch size of 16. 
FastQA \cite{son2022jointly} and Guided-QA \cite{nguyen2023emotion} were trained using the source code from each paper. Again, FastQA and Guided-QA were trained in 5 epochs with the learning rate of $5e-5$ and the batch size of 16. 

The pre-trained model \textit{TurkuNLP/wikibert-base-ja-cased} was also used for all CE models.

\begin{figure*}[!h]
  \centering
  \begin{tabular}{@{}c@{}}
    \includegraphics[height=38pt]{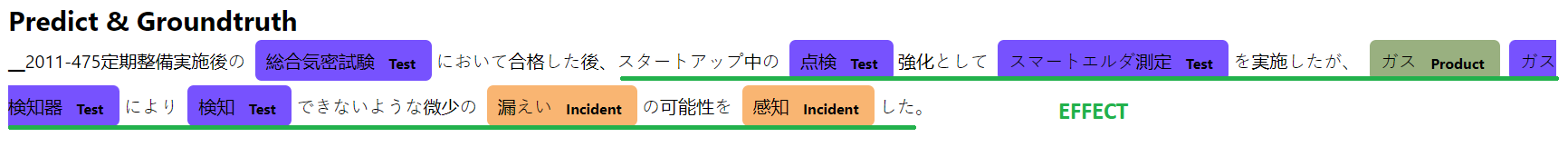}\vspace{-0.3cm} \\[\abovecaptionskip] \vspace{-0.3cm}
    \small (a) Correct Nested NER and CE example. Consecutive tokens in green denote an effect.
  \end{tabular}

  \vspace{\floatsep}

  \begin{tabular}{@{}c@{}}
    \includegraphics[height=95pt]{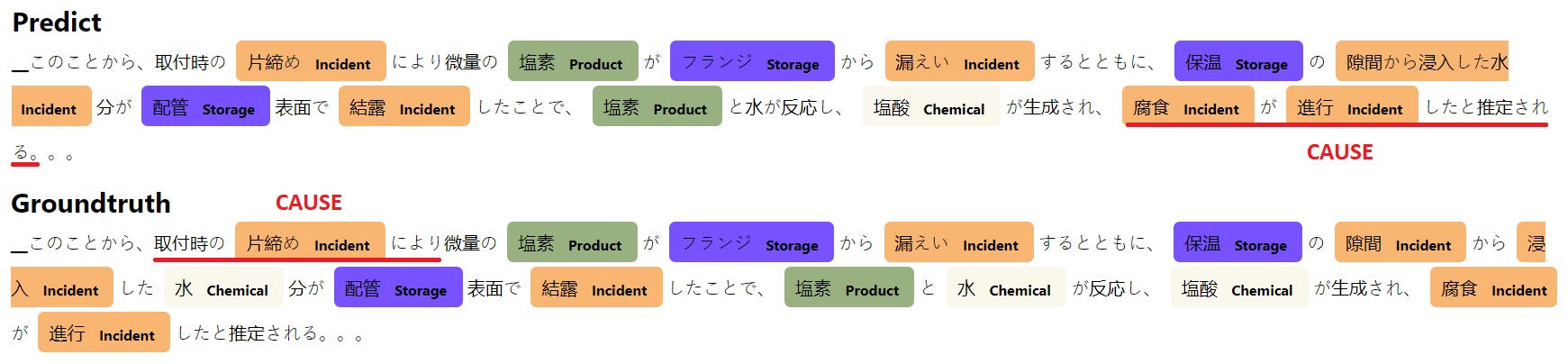} \vspace{-0.3cm} \\[\abovecaptionskip]
    \small (b) Incorrect Nested NER and CE example. Consecutive tokens in red denote a cause.
  \end{tabular}

  \caption{The figure shows a success and failure case for Nested NER and CE.}\label{fig:ner_ce_output}
\end{figure*}

\begin{figure*}[!h]
    \centering
    \includegraphics[width=0.85\textwidth]{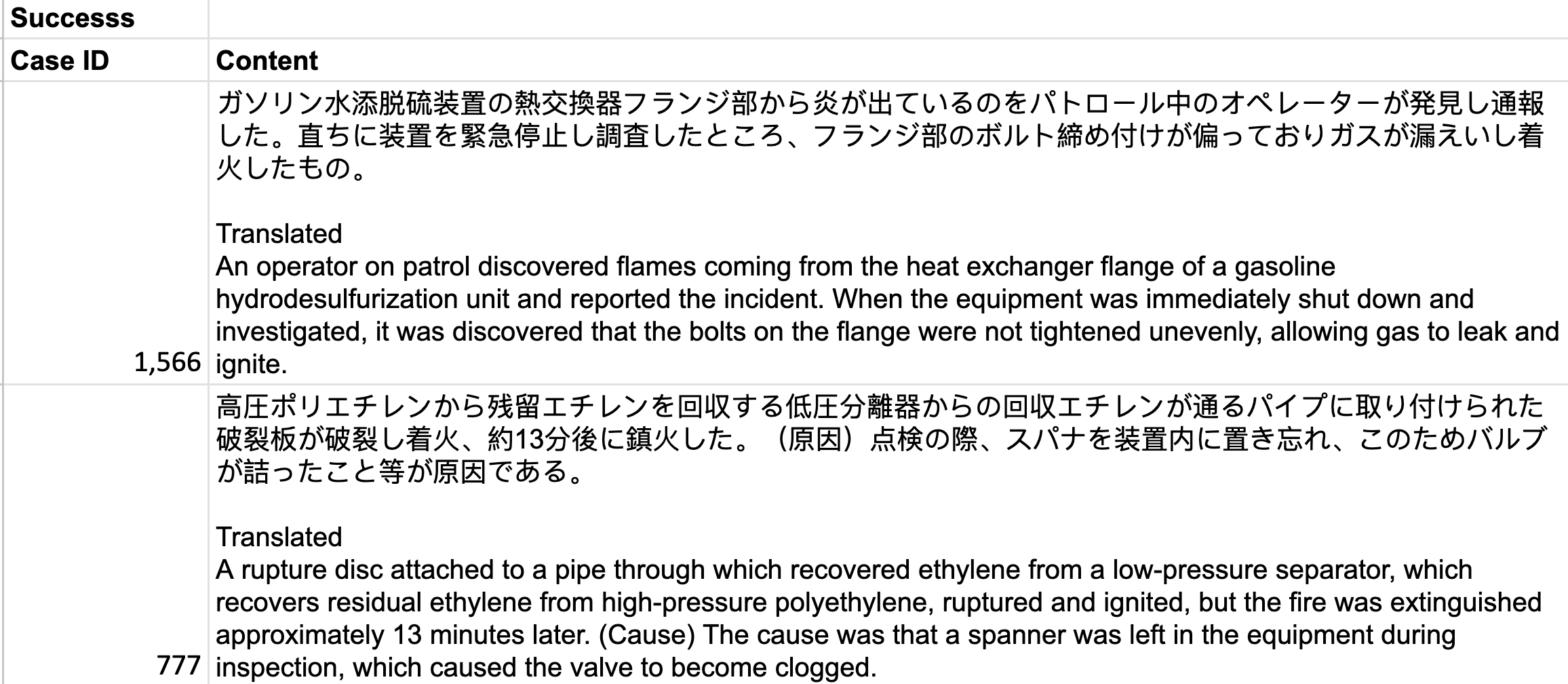}
    \caption{A sample of success case for IR model.}
    \label{fig:ir-success}
\end{figure*}

\begin{figure*}[!h]
    \centering
    \includegraphics[width=0.85\textwidth]{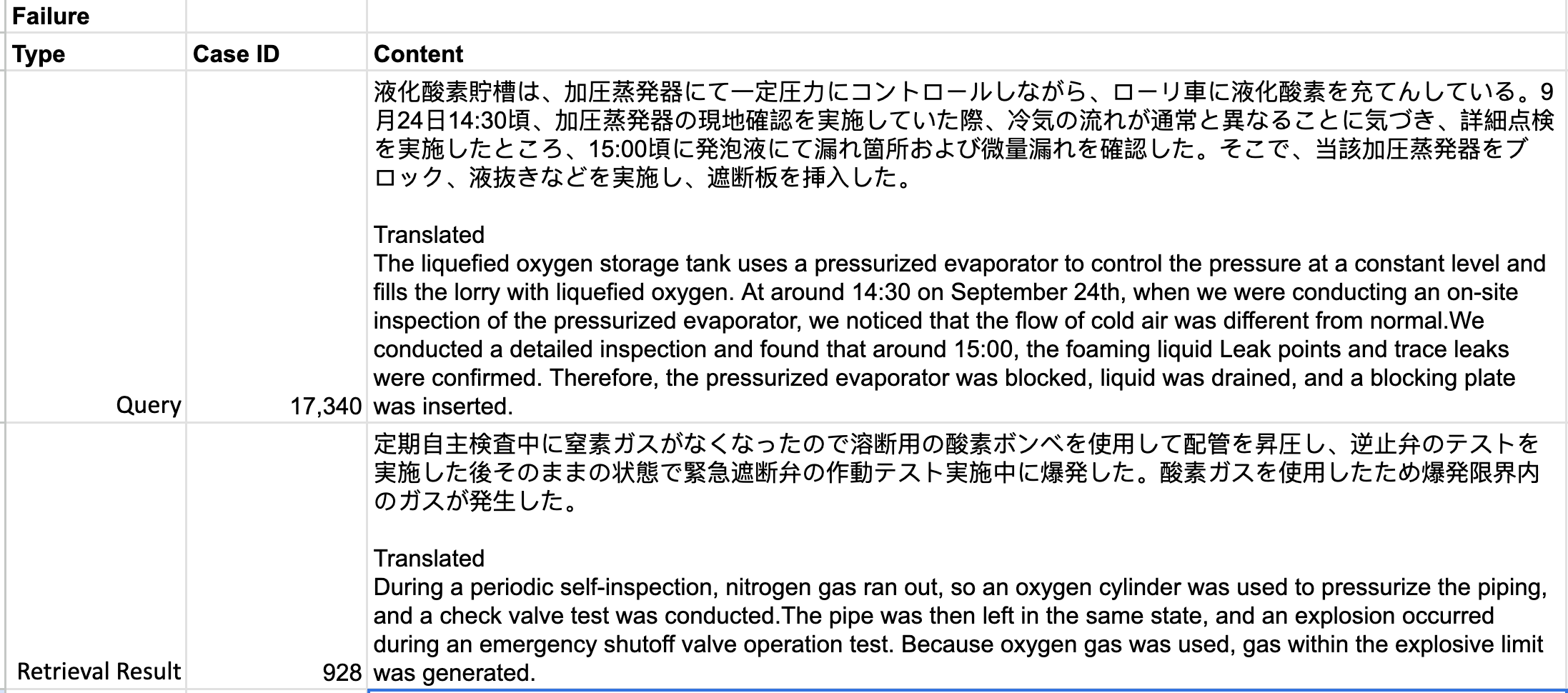}
    \caption{A failure case of IR. The retrieved sample is different in the results (leakage vs explosion).}
    \label{fig:ir-failure}
\end{figure*}

\subsubsection{IR models}\label{apx:ir}
We fine-tuned the base model \texttt{distiluse-base-multilingual-cased-v2} from sentence-BERT \footnote{https://huggingface.co/sentence-transformers/distiluse-base-multilingual-cased-v2} on the small subset of HPGIncident dataset described in Section \ref{para:IR} containing 3000 samples (not overlapped with the annotated IR dataset).

We utilize the unsupervised training objective from SimCSE \cite{gaosimcse}, which takes an input sentence and predicts itself in a contrastive objective, with standard dropout used as noise. The model is fine-tuned using 3 epochs with the learning rate of $3e-5$.
The encoder model uses mean pooling to aggregate contextual information from all tokens. We trained the model with the sequence length of 512 tokens. 

\subsection{Output observation}\label{app:output-observation}
The output of nested NER and CE is shown in Figure \ref{fig:ner_ce_output} and that of IR is shown in Figure \ref{fig:ir-success}.

\end{document}